\documentclass[10pt]{article}

\usepackage{amsmath}
\usepackage{amssymb}

\usepackage{graphicx}

\usepackage{cite}

\usepackage{color} 

\usepackage[labelfont=bf,labelsep=period,justification=raggedright]{caption}

\makeatletter
\renewcommand{\@biblabel}[1]{\quad#1.}
\makeatother

\date{}

\newcommand{\argmin}{\mathop{\mathrm{arg\,min}}} 
\newcommand{\argmax}{\mathop{\mathrm{arg\,max}}}
\newcommand{\qed}{\hfill \ensuremath{\blacksquare}}

\newif\iffig
\figtrue 

\begin{document}

\begin{flushleft}
{\Large
\textbf{Decision making under uncertainty: a quasimetric approach}
}
\\
Steve N'Guyen$^{1,\ast}$, 
Cl{\'e}ment Moulin-Frier$^{1}$, 
Jacques Droulez$^{1}$
\\
\bf{1} Laboratoire de Physiologie de la Perception et de l'Action, 
 UMR 7152 College de France - CNRS, Paris, France
\\
$\ast$ E-mail: steve.nguyen@college-de-france.fr
\end{flushleft}

\section*{Abstract}

  We propose a new approach for solving a class of discrete decision
  making problems under uncertainty with positive cost. This issue
  concerns multiple and diverse fields such as engineering, economics,
  artificial intelligence, cognitive science and many
  others. Basically, an agent has to choose a single or series of
  actions from a set of options, without knowing for sure their
  consequences. Schematically, two main approaches have been followed:
  either the agent learns which option is the correct one to choose in
  a given situation by trial and error, or the agent already has some
  knowledge on the possible consequences of his decisions; this
  knowledge being generally expressed as a conditional probability
  distribution. In the latter case, several optimal or suboptimal
  methods have been proposed to exploit this uncertain knowledge in
  various contexts. In this work, we propose following a different
  approach, based on the geometric intuition of distance. More
  precisely, we define a goal independent quasimetric structure on the
  state space, taking into account both cost function and transition
  probability.  We then compare precision and computation time with
  classical approaches.


\section*{Introduction}
It's Friday evening, and you are in a hurry to get home after a hard
day's work. Several options are available. You can hail a taxi, but
it's costly and you're worried about traffic jams, common at this
time of day. Or you might go on foot, but it's slow and tiring. Moreover, the
weather forecast predicted rain, and of course you forgot your
umbrella. In the end you decide to take the subway, but unfortunately,
you have to wait half an hour for the train at the connecting
station due to a technical incident.

Situations like this one are typical in everyday life. It is also
undoubtedly a problem encountered in logistics and control. The
initial state and the goal are known (precisely or according to a
probability distribution). The agent has to make a series of decisions
about the best transport means, taking into account both uncertainty
and cost. This is what we call \emph{optimal control under uncertainty}.

Note that he might also have an intuitive notion of some
abstract distance: how far am I from home? To what extent will it be
difficult or time consuming to take a given path? The problem might
become even more difficult if you do not know precisely what state you
are in. For instance, you might be caught in a traffic jam in a
completely unknown neighborhood.

This problem that we propose to deal with in this paper can be
viewed as sequential decision making, usually expressed as a
Markovian Decision Process (MDP) \cite{Bellman1957, Howard1960,
  Puterman1994,Boutilier1999} and its extension to Partially
Observable cases (POMDP) \cite{Drake1962,Astrom1965}. Knowing the
transition probability of switching from one state to another by
performing a particular action as well as the associated instantaneous
cost, the aim is to
define an optimal policy, either deterministic or
probabilistic, which maps the state space to the action state in order
to minimize the mean cumulative cost from the initial state to a goal
(goal-oriented MDPs).

This class of problems is usually solved by Dynamic Programming
method, using Value Iteration (VI) or Policy Iteration (PI) algorithms
and their numerous refinements.  Contrasting with this model-based
approach, various learning algorithms have also been proposed to
progressively build either a value function, a policy, or both, from
trial to trial.  Reinforcement learning is the most widely used,
especially when transition probabilities and cost function are unknown
(model-free case), but it suffers from the same tractability problem
\cite{Sutton1998}.
Moreover one significant drawback to these approaches is
that they do not take advantage of the preliminary knowledge of cost
function and transition probability.


MDPs have generated a substantial amount of work in
engineering, economics, artificial intelligence and
neuroscience, among others. Indeed, in recent years, Optimal
Feedback Control theory has become quite popular in explaining certain
aspects of human motor behavior
\cite{Todorov2002,Todorov2004}. This kind of method results in
feedback laws, which allow for closed loop control.

However, aside from certain classes of problems with a convenient
formulation, such as the Linear Quadratic case and its extensions
\cite{Stengel1986}, or through linearization of the problem, achieved
by adapting the immediate cost function \cite{Todorov2009}, the exact
total optimal solution in the discrete case is intractable due to the curse
of dimensionality \cite{Bellman1957}.

Thus, a lot of work in this field is devoted to find approximate
solutions and efficient methods for computing them.

Heuristic search methods try to speed up optimal
probabilistic planning by considering only a subset of the state space
(e.g. knowing the starting point and considering only reachable
states).
These algorithms can provide offline optimal solutions for the
considered subspace \cite{Barto1995,Hansen2001,Bonet2003}.

Monte-Carlo planning methods that doesn't manipulate probabilities
explicitly have also proven very successful for dealing with problems
with large state space \cite{Peret2004b,Kocsis2006}.

Some methods try to reduce the dimensionality of the problem in order
to avoid memory explosion by mapping the state space to a smaller
parameter space \cite{Buffet2006,Kolobov2009} or decomposing it
hierarchically \cite{Hauskrecht1998, Dietterich1998, Barry2011} .

Another family of approximation methods which has recently proven very
successful \cite{Little2007} is the ``determinization''. Indeed,
transforming the probabilistic problem to a deterministic one
optimizing another criterion allows the use of very efficient
deterministic planner  \cite{Yoon2007,Yoon2008,Teichteil-Konigsbuch2010}.

What we propose here is to do something rather different, by
considering goal-independent distances between states. To compute the distance we
propose a kind of determinization of the problem using a one step
transition "mean cost per successful attempt" criterion, which can
then be propagated by triangle inequality.  The obtained distance
function thus confers to the state space a quasi-metric structure that
can be viewed as a Value function between all states.
Theses distances can then be used to compute an offline policy using a
gradient descent like method. 

We show that in spite of being formally suboptimal (except for the
deterministic and a described particular case), this method exhibits
several good properties.
We demonstrate the convergence of the method and the possibility to
compute distances using standard deterministic shortest path algorithms.
Comparison with the optimal solution is described for different classes
of problems with a particular look at problems with \emph{prisons}.
Prisons or absorbing set of states have been recently shown to be
difficult cases for state of the art methods \cite{Kolobov2012} and we
show how our method naturally deals with these cases.

\section*{Materials and Methods}

\subsection*{Quasimetric}

Let us consider a dynamic system described by its state $x \in
X$ and $u \in U(x)$, the action applied at state $x$ leading to an associated
instantaneous cost $g(x,u)$. The dynamics can then be described by the Markov model:
\[
P(X^{t+1}|X^t,U^t)
\] 
where the state of the system is a random variable $X$ defined
by a probability distribution. Assuming stationary dynamics, a function $p\colon X^2U \to [0,1]$ exists, satisfying
\[
P([X^{t+1}=y]|[X^t=x],[U^t=u])=p(y|x,u)
\]  
This model enables us to capture uncertainties in the knowledge of the
system's dynamics, and can be used in the Markov Decision Process
(MDP) formalism.
The aim is to find the optimal policy $U(x)$ allowing a goal state to
be reached with minimum cumulative cost. 
The classic method of solving this is to use dynamic programming to
build an optimal Value function $v\colon X \to \mathbb{R}$, minimizing
the total expected cumulative cost using Bellman equation:
\begin{equation}\label{eq:stoch_bellman1}
v(x)=\min_u \left\{ g(x,u)+\sum_y v(y)p(y|x,u) \right\}
\end{equation}
which can be used to specify an optimal \emph{control
  policy}\\ $\pi\colon X\to U(X)$
\begin{equation}\label{eq:stoch_bellman2}
\pi(x) \in \argmin_u \left\{ g(x,u)+\sum_y v(y)p(y|x,u) \right\}
\end{equation}

In general this method is related to a goal state or a discount factor.

Here we propose a different approach by defining a goal
independent \emph{quasimetric} structure in the state space, defining for each state couple
a distance function $d(x,y)$ reflecting a minimum cumulative cost.

This distance has to verify the following properties:
\[
\left\{
\begin{array}{l l l}
\forall x,y\colon d(x,y) &\geq& 0\\
\forall x\colon d(x,x) &=& 0\\
\forall x,y\colon d(x,y) & =&0 \implies x=y\\
\forall x,y\colon d(x,y)& =&\displaystyle \min_z \left\{d(x,z)+d(z,y)\right\}\\
\end{array} \right.
\]
leading to the triangle inequality
\[
\forall x,y,z\colon d(x,y)\leq d(x,z)+d(z,y)
\] 

Therefore, the resulting quasi-distance function $d\colon X\times X \to \mathbb{R}^+$
confers the property of a quasimetric space to $X$.

Notice that this metric need not be symmetric (in general
$d(x,y) \neq d(y,x)$). It is in fact a somewhat natural property,
e.g. climbing stairs is (usually) harder than going down.
  
By then choosing the cost function $g(x,u)>0$ this
distance can be computed iteratively (such as the Value function).\\

For a deterministic problem, we initialize with:

\[
\left\{
\begin{array}{l l l}
d^0(x,x)&=&0\\
d^0(x,y\neq x)&=&+\infty\\
d^1(x,y)&=&\min\left\{d^0(x,y),\min\limits_{u|y=next(x,u)}\{g(x,u)\}\right\}
\end{array} \right.
\]

with $next(x,u)$ the discrete dynamic model giving the next state $y$
by applying action $u$ in state $x$.
Then we apply the recurrence: 
\begin{equation}\label{eq:qd_rec}
d^{i+1}(x,y)=\min_{z} \left\{d^i(x,z)+d^i(z,y)\right\} \quad \forall i>0
\end{equation}
We can show that this recurrence is guaranteed to converge in finite
time for a finite state-space problem.

\begin{proof}
\mbox{}
\begin{enumerate}
\item by recurrence $\forall (x,y), \forall i \colon d^i(x,y) \geq 0$ as:
\begin{itemize}
\item $\forall (x,y)\colon$\\ 
$d^1(x,y)= \min\left\{d^0(x,y),\min\limits_{u|y=next(x,u)}\{g(x,u)\}\right\}
  \geq 0$ as\\ 
$\forall (x,y) \colon  d^0(x,y)\geq 0$ and $g(x,u)>0$ by definition.
\item and if $d^i(x,y) \geq 0$ then\\ $d^{i+1}(x,y) = \min_{z}
  \left\{d^i(x,z)+d^i(z,y)\right\} \geq 0$
\end{itemize}

\item $\forall (x,y), \forall i \colon d^{i+1}(x,y) \leq d^{i}(x,y)$
  as:\\ 
$d^{i+1}(x,y)=\min_{z} \left\{d^i(x,z)+d^i(z,y)\right\}$ then
$d^{i+1}(x,y) \leq d^i(x,z)+d^i(z,y)$ in particular if we take $z=x$
we have $d^{i+1}(x,y) \leq d^{i}(x,y)$.

\item $\forall (x,y) \colon d^{i}(x,y)$ is a decreasing monotone
  sequence bounded by $0$. 
\qed
\end{enumerate}
\end{proof}

However, finding a way to initialize
$d(x,y)$ (more precisely $d^1(x,y)$) while taking uncertainty into
account, presents a difficulty in probabilistic cases as we cannot use
the cumulative expected cost like in Bellman equation.

For example we can choose:
\[
d^1(x,y)=\min\left\{d^0(x,y),\min\limits_{u}\left\{\frac{g(x,u)}{p(y|x,u)}\right\}\right\}
\]
for the first iteration with $\frac{g(x,u)}{p(y|x,u)}$ as the \emph{one-step distance}.

The quotient of cost over transition
probability is chosen as it provides an estimate of the \emph{mean cost per successful
attempt}.
If we attempt $N$ times the action $u$ in state $x$ the cost will
be $N.g(x,u)$ and the objective $y$ will be reached on average
$N.p(y|x,u)$ times.
The mean cost per successful attempt is:

\[
\frac{N.g(x,u)}{N.p(y|x,u)}=\frac{g(x,u)}{p(y|x,u)}
\]

This choice of metric is therefore simple and fairly convenient.  All the possible consequences of
actions are clearly not taken into account here, thus inducing a huge
computational gain but at the price of losing the optimality.
In fact, we are looking at the minimum over actions of the \emph{mean
  cost per successful attempt}, which can be viewed as using the best
mean cost, disregarding unsuccessful attempts, i.e. neglecting the
probability to move to an unwanted state. 

In a one-step decision, this choice is a reasonable approximation of
the optimal that takes both cost and probability into
account. 

This cost-probability quotient was used before to
determinize probabilistic dynamics and extract plans
\cite{Keyder2008,Barry2011,Kaelbling2011}.  Here we generalize this method to
construct an entire metric in the state space using triangle inequality.

We also notice that contrary to the dynamic programming approach, the
quasimetric is not linked to a specific goal but instead provides a
distance between any state pair. 
Moreover, using this formalism, the instantaneous cost function $g(x,u)$
is also totally goal independent and can
represent with greater ease any objective \emph{physical} quantity, such as
consumed energy. 
This interesting property allows for much more adaptive control since the
goal can be changed without the need to recompute at all.
As shown in the following, it is even possible to replace the goal
state by a probability distribution over states.
Another interesting property of the quasi-distance ${d\colon X\times X \to
  \mathbb{R}^+}$ is that it doesn't have local minimum from the action
point of view.

In fact, for any couple $(x,y)$, $\left\{d^0(x,y), d^1(x,y),\ldots
  d^n(x,y) \right\}$ is a decreasing finite series of non-negative numbers
(finite number of states), which therefore converges to a non-negative
number
\[
d(x,y)=\lim_{n\to \infty}\left\{ d^n(x,y)\right\}
\]\\

Note that if we multiply the cost function by any positive
constant, the quasimetric is also multiplied by the same constant.
This multiplication has no consequence on the structure of the
state space and leaves the optimal policy unchanged, therefore we can choose a constant such that:
\[
\min_{x,u}\left\{g(x,u)\right\}=1
\]

Let $D_k^n(y)$ be the subset of $X$ associated with a goal $y$ such
that:
\[
x \in D_k^n(y) \Leftrightarrow d^n(x,y)<k
\]
and let $D_k(y)=D_k^{\infty}(y)$ the subset of $X$ associated with the
goal $y$ such that:
\[
x \in D_k(y) \Leftrightarrow d(x,y)<k
\]
The subset $D_{\infty}(y)$ is the set of states from which the goal
$y$ can be reached in a finite time with a finite cost. Starting from
$x \notin D_{\infty}(y)$ the goal $y$ will never be reached either
because some step between $x$ and $y$ requires an action with an
infinite cost, or because there is a transition probability equal
to $0$.

Then the defined quasimetric admits no local minimum to a given goal
in the sense that for a given $k$, if $x \in D_k(y)$ is such that:
\[
\forall z \in D_k(y), \forall u \in U\colon P(z|x,u)>k^{-1} \text{ and
}
d(z,y)>d(x,y)
\]  
then $x=y$

\begin{proof}
\mbox{}
\begin{enumerate}
\item if $x \neq y$ and $x \in D_k^1(y)$, then $\exists u\colon
  P(y|x,u)>\frac{g(x,u)}{k}\geq k^{-1}$ and\\ $d(x,y)>d(y,y)=0$. As $y
  \in D_k(y)$ it is a counterexample of the definition.
\item if $x \neq y$ and $x \notin D_k^1(y)$, then $\exists n,z\colon
  d(x,y)=d^n(x,z)+d^n(z,y)$. As $d^n(x,z)\geq 0$, $d(z,y)\leq d^n(z,y)
  \leq d(x,y) < k$, therefore $z \in D_k(y)$. 
  \begin{itemize}
  \item If $n=1$, $\exists u\colon P(z|x,u)>\frac{g(x,u)}{k}\geq
    k^{-1}$ it is a counterexample of the definition.
  \item If $n>1$, $\exists z'\colon
    d^n(x,z)=d^{n-1}(x,z')+d^{n-1}(z',z)$. As $d^{n-1}(x,z')\geq 0$ we
    have still $d(z',y)\leq d^{n-1}(x,z')+d^{n}(z,y)\leq d(x,y)<k$ and
    therefore $z' \in D_k(y)$. 
    \begin{itemize}
    \item if $n-1=1$ it is a counterexample.
    \item else we repeat the search for intermediary state. Thus by
      recurrence, there exists some state $z_1 \in D_k(y)$ such that
      $x \in D_k^1(z_1)$ which gives a counterexample to the definition.  
    \end{itemize}
  \end{itemize}

\end{enumerate}

Consequently, if $x \in D_{\infty}(y)$, one can set $k=d(x,y)$ (a
finite distance) and apply the above property to show that there
exists at least one action $u$ transforming the state $x$ to some
state $z$ with a transition probability $P(z|x,u)>k^{-1}$ such that
$d(z,y)<d(x,y)$.
\qed
\end{proof}\\

\subsection*{HMM case}

In the real world, the state of the system is never really known. The
only available knowledge we have consists of a series of observations
reflecting \emph{hidden} states. Probabilistic inference based on
transition probability and observation likelihood allows to compute the
probability distribution over the hidden states.  This class of
systems is usually modeled as Hidden Markov Models (HMM) and the
problem of controlling such a system becomes a Partially Observable
Markov Decision Process (POMDP).

Extending the quasimetric method to the POMDP case
does not however, come without cost.
Ideally, as with the theoretical POMDP, we should define a quasi-distance
not just on the state space but on the belief space (estimated distribution
over states), which is continuous and consequently difficult to deal with
\cite{Kaelbling1998}.

A possible approximation is to compute the policy not on the
belief space, but on the observations-actions space, obtaining
$P([U^t=u]|o_{0:t},u_{0:t-1})$.

Let us assume that we have a state observer maintaining a distribution over
states, knowing all previous observations and actions $P([X^t=x]|o_{0:t},u_{0:t-1})$.

At time $t$ we know all the observations $o_{0:t}$ and all the
previous actions $u_{o:t-1}$, thus the distribution for the state can
be recursively updated by the forward HMM equations:
\begin{align*}
P([X^t=x]|o_{0:t},u_{o:t-1})&\propto P([O^t=o]|[X^t=x])\\ &\times
\sum_{y} \big [P([X^t=x]|[X^{t-1}=y],u^{t-1})\\
&\times P([X^{t-1}=y]|o_{0:t-1},u_{0:t-2}) \big ]
\end{align*}
 
with $P(O^t|X^t)$ the observation model.

Then the distribution over action space can be computed by
marginalizing over state space:
\begin{align}\label{eq:pomdp}
\nonumber P([U^t=u]|o_{0:t},u_{0:t-1})=&
\sum_x \big [P([X^t=x]|o_{0:t},u_{0:t-1})\\ &\times P([U^t=u]|[X^t=x])
\big ]
\end{align}
assuming we have already computed the state dependent action policy $P([U^t=u]|[X^t=x])$ (see below).

Following this, a decision must be made based on this distribution. The chosen action can be random
\[
u_t^{random}\sim P([U^t=u]|o_{0:t},u_{0:t-1})
\]
the most probable
\[
u_t^{max}= \argmax_u P([U^t=u]|o_{0:t},u_{0:t-1})
\]
or the \emph{mean} 
\[
u_t^{mean}= \sum_u u.P([U^t=u]|o_{0:t},u_{0:t-1})
\]

Here we assume a separation between state estimation and control,
considerably reducing the computational cost compared to the optimal
POMDP solution, which is intractable for most real-life problems.

One drawback however, is that the resulting policy could be less
optimal and lacking in information-gathering behavior, for example.

\subsection*{Probabilistic policy}

As we have seen, in the classic MDP formalism, the policy $\pi(x)$ is
a \emph{deterministic} mapping of the state space $X$ toward the
action space $U$ (using $\displaystyle \argmin_u$). Pure MDP formalism only
considers the optimal action (greedy policy), so a choice is made
during the computation of the policy to only consider the one action
that minimizes the cost.

However, this method could be viewed as arbitrary to a certain extent, especially for
multimodal cases where the choice of a unique optimal
action may lead to loss of information or blocking behavior. 

In the field of reinforcement learning, the greedy policy is usually
avoided in order to maintain exploratory behavior.
To do so, methods
such as $\epsilon$-greedy, soft $\epsilon$-greedy and soft-max action
selection were employed \cite{Sutton1998}.

Here we propose building $P(U^t|X^t, Y^t)$
with $Y$ the goal, using a Gibbs distribution (soft-max like form):

\begin{equation}\label{eq:smpolicy}
P([U^t=u]|[X^t=x],[Y^t=y]) = \frac{e^{-\beta D_u d(x,y)}}{\displaystyle \sum_{u_i} e^{-\beta D_{u_i} d(x,y)}}
\end{equation}
with $\beta$ a parameter modulating the \emph{sharpness} of the
distribution (and consequently the exploration rate), and $D_u d(x,y)$ a \emph{probabilistic gradient} of
the quasi-distance:

\begin{equation}\label{eq:probgrad}
D_ud(x,y)=g(x,u)+\sum_zd(z,y)p(z|x,u)-d(x,y)
\end{equation}

This gradient takes the immediate cost of the action into account, as well as the
difference between the expected and current quasi-distances.

The resulting distribution depends on a goal $y$, which can be
fixed or even an evolving distribution $P(Y^t)$. The latter
distribution can represent multiple objectives or just uncertainty
with respect to the goal. We can then obtain a the state dependent
action policy by marginalizing:

\begin{equation}\label{eq:probgoal}
P(U^t|X^t)=\sum_y P(U^t|X^t,Y^t).P(Y^t)
\end{equation}

This way to build a policy can certainly be applied to any \emph{potential}, such as
the Bellman Value function.
Similarly to reinforcement learning methods, actions are weighted
according to their ``value estimate'' which, in our case, is the
gradient of the expected quasi-distance.
In MDP, the current state is known, so that the probability
distribution over action space is directly given by the state
dependent action
policy (eq.~\ref{eq:smpolicy} or eq.~\ref{eq:probgoal}).
In POMDP, the current state is not known, but by marginalization over
the state space, one can also compute the distribution over the action
space (eq.~\ref{eq:pomdp}).

We can see that if $\beta \to \infty$, the policy tends toward a Dirac delta
distribution (if a unique action minimizing the value exists). 
This extreme case reduces to the MDP optimal policy where a unique
optimal action is mapped to each state, similarly to the Value
Iteration or Policy Iteration methods.

The knowledge of a distribution on $U$ allows a random draw decision
to be made from the distribution which can be useful to avoid blocking
behavior or even learning.  According to the $\beta$ value, the
soft-max policy associated with the random draw decision generates
either a more optimal behavior (large $\beta$) or a more exploratory
behavior (small $\beta$).

\section*{Results and discussion}
\subsection*{Comparison with dynamic programming}

\subsubsection*{Convergence and complexity}  

First as we have shown, computation of the quasi-distance is ensured to
converge even for infinite horizon (in finite time for a finite state
space) while the standard Value Iteration algorithm is not. In fact,
it is usually necessary to introduce
a \emph{discount factor} $\gamma$ in Bellman's equation to assure
convergence, but at the price of sub-optimality whereas such a thing
is not needed with the quasimetric.

Constructing the initial \emph{one-step distance} $d^1(x,y)$ for all
state couples is in $O(|U|.|X|^2)$.
Then, directly applying the recurrence in equation \ref{eq:qd_rec}
leads to a complexity of $O(|X|^3.log|X|)$ for the whole state space
(all-to-all states).
However, the quasimetric construction uses probabilities only at the
first iteration (i.e. the \emph{one-step distance}) and then propagates
these distances with triangle inequality.
This propagation of \emph{one-step distances} is completely
deterministic and no more probabilities appear afterward.
Thus, computing the quasi-distance can be reformulated
in a graph theory framework as a deterministic shortest path problem.

Let us consider the weighted directed graph (or network) $G=(V,A)$ with
vertices $V=X$ and arcs $A$ the set of ordered pairs of vertices.
We can assign to each oriented arc $e=(x,y)$ the weight
$w_{x,y}=d^1(x,y)$ (the \emph{one-step distance}). 
Remark that for the sake of efficiency it is preferable to consider an arc
only if the associated weight $w_{x,y} \neq +\infty$, i.e. if an
action $u$ exists with a finite cost and for which the transition probability $p(y|x,u) \neq 0$.
Constructing this graph is of the same complexity than the $d^1(x,y)$
iteration and is only computed once nonetheless.

Then, the problem of computing the quasi-distance from $x \to y$
becomes the problem of finding the length of the shortest path between vertices $x$
and $y$.

One can compute the whole quasimetric (all-to-all states) by computing
the all-pairs shortest paths using for example the
Floyd-Warshall algorithm \cite{Roy1959,Floyd1962,Warshall1962}.  However,
considering the usual MDP problem with a fixed goal, one would
prefer to compute the quasi-distance for only one goal, which can be viewed
as the multiple-source shortest path. An efficient way to solve this
is to consider the transposed graph $G^T$ in which arcs are inverted
and to solve the single-source shortest path (from goal vertex) using
for example Dijkstra's algorithm \cite{Dijkstra1959} or $A^*$
depending on the problem \cite{Hart1968}.

From a computational point of view, using Dijkstra's
algorithm to solve the one-goal problem can be done with the $O(|A| +
|V|.log|V|)$ worst case complexity using the appropriate data structure \cite{Fredman1987}. Knowing that $|V|=|X|$ and
$|A|\leq |X|^2$ it's $O(|X|^2 + |X|.log|X|)=O(|X|^2)$, considering a
fully connected graph.

This is to be compared to classical discounted Value Iteration method
which complexity is $O(|U|.|X|^2)$ for one iteration (or sweep) with
the worst case number of iterations to converge proportional to
$\frac{1}{1-\gamma}log(\frac{1}{1-\gamma})$, $\gamma$ being the discount factor \cite{Littman1995}.

Notice that, transition probabilities are usually sparse allowing
the graph to be equally sparse. Hence, considering the mean vertex
out-degree $\hat{D}^{+}$, complexity using Dijkstra's method becomes
$O(\hat{D}^{+}.|X|+|X|.log|X|)$, $\hat{D}^{+}$ depending on the
dispersion of transition probabilities. 

Therefore, quasi-distance can then be easily solved in a
computationally efficient way using the standard deterministic graph
theory methods.

\subsubsection*{Equivalence}

The question is, how much does the quasimetric method diverge from
the dynamic programming? 
In other words, how can we compare the quasi-distance with 
the value function in order to discuss the optimality approximation?
In order to be able to compare, we first have to consider only a subset
of the quasimetric by looking at the quasi-distances from all states to one
unique state (a goal).
If the quasi-distance and the value function are equal for a specific goal (strong
equivalence) then clearly policies obtained with both methods will
lead to the same behavior. But it is also possible that the
quasi-distance differs from the value function and still yield the same
policy (weak equivalence).

In the deterministic case, the quasi-distance and the value function
are trivially equal. 
But there is at least one other class of problems where these two
approaches are strictly equivalent that we call the \emph{probabilistic maze}.

Let us consider a probabilistic system where the uncertainties concern
the success of actions. If an action $u$ succeeds, it
drives the system from one state $x$ to another state $next(x,u)$; if it
fails, the system remains in state $x$.  We can then call
$p_{suc}(u,x)$ the probability that action $u$ is successful from
state $x$. This function determines all the transition probabilities
that are null except for:

\begin{equation}\label{eq:probmazepsuc}
\left\{
\begin{array}{l l l}
P(next(x,u)|x,u)&=&p_{suc}(u,x)\\
P(x|x,u)&=&1-p_{suc}(u,x)
\end{array} \right.
\end{equation}

This kind of systems was also described as ``self-loop MDPs'' and used
for MDPs determinization \cite{Keyder2008}.
For this class of systems -- which includes those that are deterministic --
the value function and the quasi-distance are strictly equivalent and
lead to the same optimal policy.

Indeed, we can inject these probabilities in the Bellman's equation:
\begin{align}\label{eq:probmazebellman}
\nonumber V(x)=&\min_u\big\{ g(x,u)+V(x)(1-p_{suc}(u,x))\\ 
&+V(next(x,u))p_{suc}(u,x)\big\}
\end{align}
So
\begin{equation}\label{eq:probmazebellman2}
\min_u\big\{ g(x,u)-p_{suc}(u,x)(V(x)-V(next(x,u))\big\}=0
\end{equation}

In each state $x$ there exists at least one optimal action $u^*(x)$
such that:
\begin{align}\label{eq:probmazeuopt}
\nonumber u^*(x) \in \argmin_u\big\{& g(x,u)+V(x)(1-p_{suc}(u,x))\\
&+V(next(x,u))p_{suc}(u,x)\big\}
\end{align}

In a probabilistic maze, an action can only succeed -- thus driving the
system from $x$ to $next(x,u)$ -- or fail and leaving the system
unchanged.
So starting from any initial state $x_0$, the optimal policy $u^*$
describes a unique optimal trajectory $\{x_0 \to x_1 \to \dots \to
x_n=goal\}$. 
If the optimal action fails in some state $x_i$,
the system remains in $x_i$ and the optimal action to apply is still
the same. So $u^*(x_i)$ is repeatedly chosen until it succeeds to move
the system to $x_{i+1}=next(u^*(x_i),x_i)$ and the probability to
succeed in exactly $k$ tries is $p_{suc}(u,x).(1-p_{suc}(u,x))^{k-1}$.
Therefore, the mean cost for the transition $x_i \to
x_{i+1}$ is:

\begin{align}\label{eq:probmazemeancost}
\nonumber g(x_i \to x_{i+1})=&p_{suc}(u,x).g(u^*(x_i),x_i)\\
&\times \sum_{k=1}^{\infty}k.(1-p_{suc}(u,x))^{k-1}
\end{align}
and as $\displaystyle \sum_{k=1}^{\infty}k.(1-p_{suc}(u,x))^{k-1}=\frac{1}{p_{suc}(u,x)^2}$ we have:
\begin{equation}\label{eq:probmazemeancost2}
g(x_i \to x_{i+1})=\frac{g(u^*(x_i),x_i)}{p_{suc}(u,x)}
\end{equation}

So the optimal policy is the one which minimizes this mean cost per
successful attempt
\begin{equation}\label{eq:probmazeu}
u^*(x_i) \in \argmin_u\bigg \{\frac{g(u^*(x_i),x_i)}{p_{suc}(u,x)}\bigg\}
\end{equation}
Finally we have:
\begin{equation}\label{eq:probmazevd}
V(x_0)=\sum_{i=0}^{n-1}\min_u \bigg \{
\frac{g(u(x_i),xi)}{p_{suc}(u(x_i),x_i)} \bigg \} = d(x_0,x_n)
\end{equation}\\

\begin{figure}[ht]
\iffig
\begin{center}
\includegraphics[width=4in]{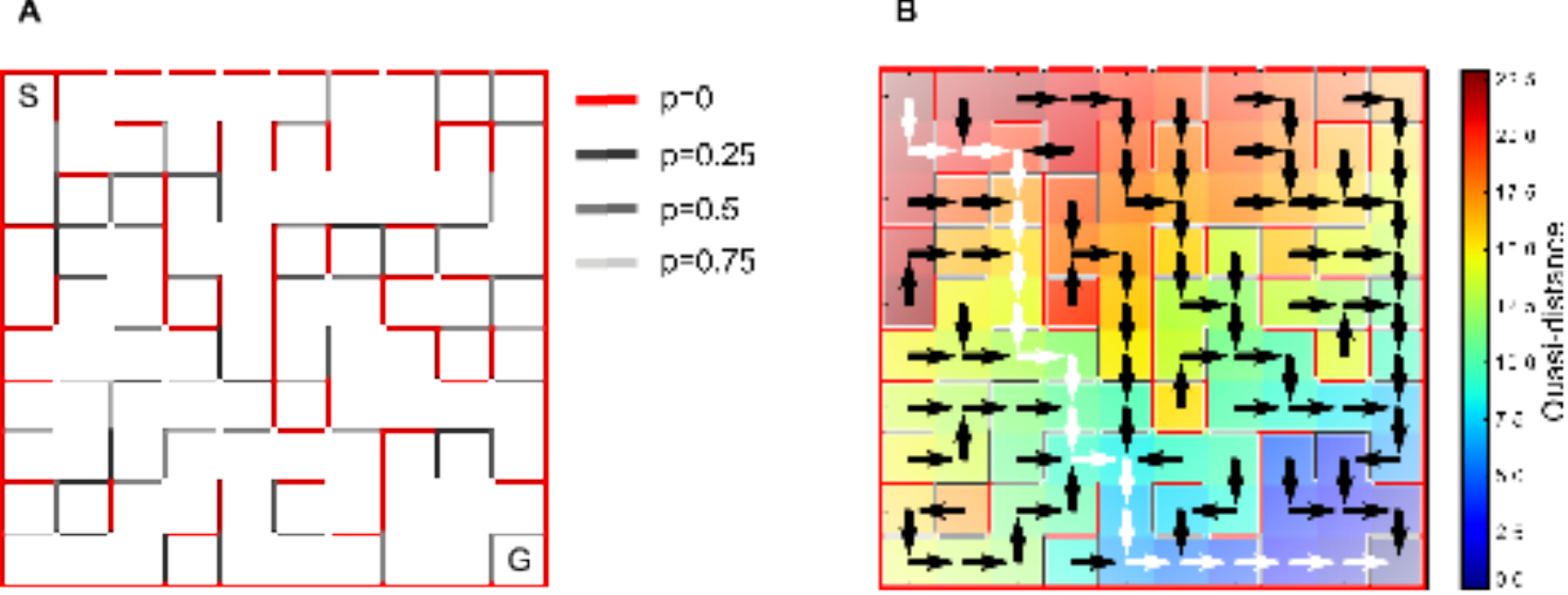}
\end{center}
\fi

\caption{{\bf Example of a simple \emph{probabilistic} maze of size
  $10\times 10$ where dynamic programming and quasimetric methods are
  equivalent.} (A) S is the starting state and G the goal. The red wall
  cannot be traversed (transition probability $p=0$) while gray ones
  can be considered as \emph{probabilistic doors} with transition
  probability $0 < p < 1$. 5 actions are considered: not moving, going
  east, west, south and north. (B) Quasi-distance obtained for the
  probabilistic maze example with a constant cost function ($\forall
  x,u\colon g(x,u)=1$) and corresponding policy. White arrows
  represent the optimal policy from position S to G. Black arrows
  represent the optimal policy to reach G from other positions.}
\label{fig:laby}
\end{figure}

Figure \ref{fig:laby} shows an example of such a maze, with the corresponding quasi-distance and policy. 

This type of problem may appear somewhat artificial but it can for
example, refer to a \emph{compressed} modeling of a deterministic
system, exploiting the \emph{structure} of the state space.

Let us imagine a mobile agent in a corridor. In a discretized space,
the corridor can have length $n$, each action moving the mobile one
cell forward with the probability $p \simeq 1$.

In order to exit the corridor, the action has to be
applied $n$ times. Alternatively, this discrete space can be \emph{compressed} by representing the corridor with a single cell and the
probability to succeed (i.e. to exit the corridor)
$p=\frac{1}{n}$. The resulting model is the probabilistic maze
described above.

\subsubsection*{Non-equivalence}

In the general case, the quasimetric approach will differ
from the dynamic programming method. These differences arise when the transition
probabilities are \emph{spread out} along several arrival states. This
dispersion of arrival states can produce differences between the
quasi-distance and the value function with -- or without -- differences
in the optimal policy obtained.

\paragraph*{Systems yielding a quasi-distance different to the
  value function}

\begin{figure}[ht]
\iffig
\begin{center}
\includegraphics[width=4in]{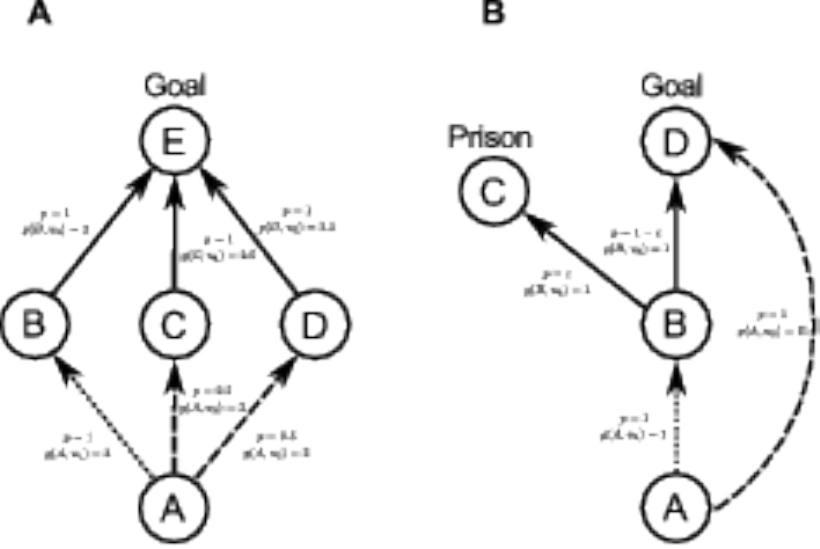}
\end{center}
\fi

\caption{{\bf Simple systems where the quasimetric and the dynamic
    programming methods are not equivalent. }(A) Example of non deterministic systems where the
  quasi-distance differs from the value function.Arrows indicate
  possible actions with their associated transition probabilities $p$
  and costs. Dotted arrow represents action $u_1$ and dashed arrows
  action $u_2$, both allowed in state $A$.
  (B) Example with a prison state $C$. Starting from $A$ to the goal $D$
  we can choose between two actions. Action $u_1$ in dotted leads to $B$
  with a low cost but then with the risk to fall from $B$ to $C$
  with a probability $p=\varepsilon$. $B$ is a risky
  state. Action $u_2$ in dashed leads to the goal with a probability
  $p=1$ but with a high cost $\Omega$.}
\label{fig:exemples}
\end{figure}

Here is a simple case illustrating the difference between the two
methods.
Let us consider a system with fives states $\{A,B,C,D,E\}$ where $A$ is
the starting state and $E$ the goal (cf. Fig.~\ref{fig:exemples}A). 
This system is almost deterministic since the only uncertainty
relates to one action in state $A$. For $A$ there are two possible
actions, one driving the transition $A \to B$ with a probability of $1$ and a cost of
$3$ (action $u_1$) and one driving either $A \to C$ or $A \to D$ with a probability of $0.5$
and a cost of $2$ ($u_2$). 
Then from $B$, $C$ and $D$ the transition are deterministic, with
associated costs of respectively $2$, $2.5$ and $2.5$. 

The corresponding computed quasi-distances can be found in table~\ref{tab:ex1_qd_v}.
The shortest path according to the quasi-distance is
$A \to B \to E$. The optimal policy in $A$ however, is to choose the
action $u_2$ leading to either $C$ or $D$ with a probability of 0.5 and a
cost of 2. Indeed, for the action $u_1$ leading to $B$ we have $g(A,u_1)+V(B)=5$ whereas $g(A,u_2)+0.5.V(C)+0.5.V(D)=4.5$.

The value function of state $A$ is slightly lower than $d(A,E)$ but
both methods lead to the same optimal choice of
$u_2$ while in $A$.

In this example, the quasi-distance yields an inaccurate estimate
of the mean cost when starting from state $A$. In fact, the quasi-distance
computation tends to favor actions with low dispersion in transition
probabilities (low uncertainty).
So here, the quasi-distance obtained differs from the Value function for
state $A$, but generates the same optimal policy. 

The policy can also differ in the general case.
In fact, replacing all the costs in the same example with
$1$ leads to $V(B)=V(C)=V(D)=d(B,E)=d(C,E)=d(D,E)=1$.
However, due to the uncertainty of action $u_2$ we have
$d(A,B)=d(A,C)=\frac{1}{0.5}=2$ and $d(A,D)=1$, thus clearly biasing
the policy obtained with the quasi-distance in favor of $u_1$. On
 the contrary, as the Value function takes all of the consequences
of actions into account, $u_2$ leads to $1+0.5V(B)+0.5V(C)=2$ and $u_1$ to
$1+V(D)=2$, so the two actions are equivalent.   
Roughly speaking, the quasi-distance yields an uncertainty
aversive policy, resulting from the $\frac{g}{p}$ form of the one-step distance.

\paragraph*{Systems with \emph{prison-like} states}

Control under uncertainty can be viewed as a continuous decision making
where both cost and uncertainty must be dealt with. The trade-off
between cost and uncertainty can be illustrated by the spider problem
\cite{Kappen2005} where an agent can reach a goal quickly by crossing
a narrow bridge or by slowly walking around a lake. In a deterministic
case, crossing the bridge is the obvious optimal action, but when
there is uncertainty as to whether the spider is able to cross the
narrow bridge, the optimal action could be to walk safely around the
lake (as falling into the water may be fatal).

As regards the spider problem, falling into the water may bear a
sufficiently large cumulative cost to justify choosing to walk around the lake
hazard-free.  But then, what happens when confronted with a choice
between a very costly but certain action and a low-cost action where
there is a small probability of death?
Clearly this problem may be much more difficult as death may not
be associated with a high cost per se. The action of ``walking'' when in state
``bridge'' has no objective reason to be higher than ``walking'' when
in state ``lakeside'' if we consider energy consumption.
Instead, the problem with death does not lie in the cost but in the
fact that it is an irreversible state.

In our modeling, death can be represented as a \emph{prison} state
from which one cannot escape. So this singularity is slightly
different from a state-action with a high cost.  In general these
prisons can be a subset of the state space rather than an unique
state.  These particular states, sometimes referred as ``dead-ends'',
are known to be problematic for MDPs and are implicitly excluded from
the standard definition as their existence may prevent solution to
converge \cite{Bertsekas1995}.  Moreover, recent works in the domain
of planning have identified these problems as interesting and
difficult, recognizing the need to find specific methods to deal with it
\cite{Little2007}. It is to be noted that a prison is an absorbing set
of states that are not necessarily absorbing as it may be possible to
``move'' inside the prison.
In fact a prison is as a set of states that does not contain the goal
and from which we cannot reach the goal. 

To illustrate this class of problems, let us imagine that the spider
is unable to swim. Falling into the water now leads to a prison state (death
of the spider).

Figure \ref{fig:exemples}B models this problem considering four
states $\{A,B,C,D\}$ where $A$ is the initial state and $D$ the goal.
There is a choice of actions in state $A$.
The first action $u_1$ drives the transition $A \to B$ with a
probability $p=1$ and a cost of $1$. Then from $B$, the unique action
can lead to $D$ with $p=1-\varepsilon$ or $C$ with a probability
$p=\varepsilon$ and a cost of $1$. The second action $u_2$ allows for a
transition $A \to D$ with a probability $p=1$ but a cost $\Omega$.

In this case what should the spider do?  The computed value function
and the quasi-distance can be found in table~\ref{tab:exfrogger_qd_val}.
According to Bellman's equations, the action $u_1$ should never be
attempted.  Indeed, for state $A$ and action $u_1$ we have
$g(A,u_1)+p(B|A,u_1)V(B)=+\infty$ and for $u_2$ we have
$g(A,u_2)+p(D|A,u_2)V(D)=\Omega$. So clearly the optimal choice here
according to the value function is $u_2$, independently of cost
$\Omega$ which can be seen as rather radical.

In contrast, the action policy obtained with the quasi-distance depends
on the relative values of $\Omega$ and $\varepsilon$ (cost
vs. uncertainty) that are parameters of the problem.
Indeed,
$d(A,D)=\min\{1+\frac{1}{1-\varepsilon},\Omega\}$ involves:

\[
\left\{
\begin{array}{l l l}
Du_1(A)&=&1+\frac{1}{1-\varepsilon}-\min\{1+\frac{1}{1-\varepsilon},\Omega \}\\
Du_2(A)&=&\Omega-\min\{1+\frac{1}{1-\varepsilon},\Omega \}
\end{array} \right.
\]

So if $1+\frac{1}{1-\varepsilon} < \Omega$ we have $Du_1(A)=0$ and
$Du_2(A)=\Omega-1-\frac{1}{1-\varepsilon}>0$ then $u_1$ is chosen.

But if $1+\frac{1}{1-\varepsilon} > \Omega$,
$Du_1(A)=1+\frac{1}{1-\varepsilon}-\Omega>0$ and $Du_2(A)=0$, then
$u_2$ is chosen.

We see that different policies can be chosen depending on the problem whereas
dynamic programming will always avoid $u_1$. It is then also possible
to modify the cost function in order to move the cursor of the risky
behavior by changing the cost of $u_2$.  
\newline

We can formalize these \emph{prison-like} states further in order
to better control these effects.

Let us define state $x$ as belonging to the prison $J(y)$ of state $y$
if there is no policy allowing the transition from $x$ to $y$ with a non zero
probability. 

We notice that if we compute the reaching set $Q(y)=\{x \in X :
\exists \; path(x \to y)\}$, we can obtain
$J(y)=X-Q(y)$. Then, by definition $\forall x \in J(y), \;
d(x,y)=\infty$.

With our method, these prison states are states for which the
quasi-distance to a specific (goal) state is infinite. Moreover,
contrary to dynamic programming methods, for a
finite cost function (and in a finite state space) the prison states
are the only states with infinite quasi-distance to the goal, making them easy to
identify.
In fact, as described, we initialize all distances with:
\[
d^1(x,y)=\min\left\{d^0(x,y),\min\limits_{u}\left\{\frac{g(x,u)}{p(y|x,u)}\right\}\right\}
\]
So any ``one-step'' distance between two states $x$ and $y$ will be finite if
at least one action $u$ with a non zero probability $p(y|x,u)$
exists. Then, these ``one-step'' distances are propagated by triangle
inequality ensuring that $d^{i+1}(x,y)=\min_{z}
\left\{d^i(x,z)+d^i(z,y)\right\}$ is infinite iff the probability of
reaching $y$ from $x$ is zero, i.e. there is no path between $x$ and
$y$.
Thus with our method, considering a finite cost function and a finite
state space, every prison state has an infinite distance and
every state with infinite distance is a prison state.

As described, the proposed general quasi-metric iterative algorithm
can detect all the possible prison states and for a goal directed MDP,
the proposed deterministic shortest path algorithm for computing the
quasi-distance will also naturally detect these prisons without
propagating to other states.

Furthermore, there are also \emph{risky} states that do not belong to $J(y)$ but
are still associated with an infinite Value function. Obviously, all
states in $J(y)$ have an infinite Value but contrary to the
quasimetric the reciprocal is not
true. Therefore, all states with a non zero probability of leading to a prison
state also have an infinite Value (propagated by the conditional
expectation in Bellman's equation):
\[
z \in J(y) \Rightarrow V(z)=\infty
\]
so we have

\begin{equation*}
\forall u\ \exists z \in J(y)\colon p(z|x,u) > 0 \Rightarrow \forall
u\ \sum_z V(z).p(z|x,u) = \infty \Rightarrow V(x)=\infty
\end{equation*}
Thus the infinite value can propagate to the whole state space depending on the
distributions.  This property of the dynamic programming method makes
prison states indistinguishable from other risky states if there
is no ``complete proper policy'' (a policy leading to the goal with a
probability of 1). This may also prevent any policy to be computed.  Indeed,
all the possible policy can appear to be equivalent when all states
have an infinite Value.
This case appears if we remove the action $u_2$ in this example, then
the prison becomes ``unavoidable''.
In recent years, a number of work have been devoted to
formalizing, detecting and dealing with these prisons in the planning
domain, in particular for ``unavoidable'' ones
\cite{Little2007,Kolobov2010,Teichteil-Konigsbuch2011,Kolobov2012}.
The straightforward method we propose here allows for a finer
grained management of this risk.

The set $K(y)$ of these risky states can be constructed iteratively or
by looking at $N^-(J(y))$, the set of predecessors of $J(y)$.
We can observe that $N^-(J(y))$ is the set of states for which at least
one action leads to a prison state with $p>0$.
We call this set the \emph{weakly risky} states $K'(y)$: 
\[
K'(y)=\{ x \notin J(y) \ | \ \exists u \ \exists z \in J(y)\colon
p(z|x,u)>0 \} = N^-(J(y)).
\]

The risky set is:
\[
K(y)=\{ x \in N^-(J(y)) \ | \ \forall u \ \exists z \in J(y)\colon p(z|x,u)>0 \}.
\]

We can even decide a minimal acceptable risk $\varepsilon$
($\varepsilon$-risky set), such as:
\[
K_{\varepsilon}(y)=\{ x \in N^-(J(y)) \ | \ \forall u \ \exists z \in J(y)\colon p(z|x,u)>\varepsilon \}
\]

As seen in the previous example, from a quasi-distance point of
view we can consider a risky state
to be very close to the objective. If in a particular state $x$ all actions carry the probability
$p>\varepsilon$ of entering a prison state, but at least one (let's say
$u$) also has $p>0$ of going directly to the objective $y$, we have
$d(x,y) \leq \frac{g(x,u)}{p}$.
In order to avoid these risky states it can be decided that
$\forall x \in K(y)\colon d(x,y)=\Omega$ with $\Omega$ an arbitrary
large value (possibly infinite).

This ability to deal with risk contrasts with the classic dynamic
programming method, according to which one should \emph{never} cross
the road or use a car, considering that there is always a non zero
probability of an unavoidable and irremediable accident (prison
state).  However, by crossing the road or using a car we put ourselves
in a risky state, but not in a prison state!

Consequently, a distinction between $J(y)$ and $K(y)$ along with the
ability to parametrize the risk/cost trade-off enabled by the quasimetric
approach is essential and may be interesting for modeling human
behavior.

\subsection*{Applications}

\subsubsection*{Under-actuated pendulum}

Let us consider an under-actuated pendulum driven by the following
equation:

\begin{equation}\label{eq:pendulum}
mr^2\ddot{\theta}=C+mgr.sin(\theta)
\end{equation}
with $m$ the mass, $r$ the radius, $C$ the torque, $\theta$ the
angular position and $g=9.81\ m.s^{-2}$.

The problem is to reach and maintain the unstable equilibrium
$\theta=0$ (upward vertical) from the starting stable one $\theta=\pi$
(downward vertical) with a minimum cumulative
cost, knowing that we can only apply a torque $C<mgr$.
If we use as the time unit $\tau=\sqrt{\frac{r}{g}}$, the time constant of the
pendulum, we can reduce equation \ref{eq:pendulum} to the dimensionless
one:

\begin{equation}\label{eq:pendulum_dless}
\ddot{\theta}=u+sin(\theta)
\end{equation}
with the normalized torque $u=\frac{C}{mgr}$ such as $|u| \leq u_{max} < 1$.

Then, by considering $X$, $Y$ and $U$ as the discrete variables, representing
respectively $\theta$, $\dot{\theta}$ and $u$, we can decompose the
probabilities as follows:

\begin{equation}\label{eq:pendulum_probs}
 P(X^{t+\Delta t},Y^{t+\Delta t}|X^t,Y^t,U^t)=P(X^{t+\Delta
  t}|X^t,Y^t,U^t) \times P(Y^{t+\Delta t}|X^t,Y^t,U^t)
\end{equation}

and the following discrete Gaussian forms:
\begin{equation}\label{eq:pendulum_xy}
\left\{
\begin{array}{l l l}
P(X^{t+\Delta t}|[X^t=x],[Y^t=y],[U^t=u])&\propto&\mathcal{N}(\mu_x=x+\Delta t.y
+\frac{1}{2}\Delta t^2(u+sin(\theta)),\sigma_x)\\
P(Y^{t+\Delta t}|X^t,Y^t,U^t)&\propto&\mathcal{N}(\mu_y=y+\Delta t \times (u+sin(\theta)),\sigma_y)
\end{array} \right.
\end{equation}



with $\Delta t$ the discrete time step.

So here, we approximate in discrete time the equations of the dynamics with a
Gaussian uncertainty hypothesis, described by parameters $\sigma_x$ and
$\sigma_y$. 

Simulations were done for a state space of $|X|=|Y|=51$ and
$|U|=21$ with $\sigma_x=\sigma_y=0.2$.

\begin{figure}[ht]

\iffig
\begin{center}
\includegraphics[width=4in]{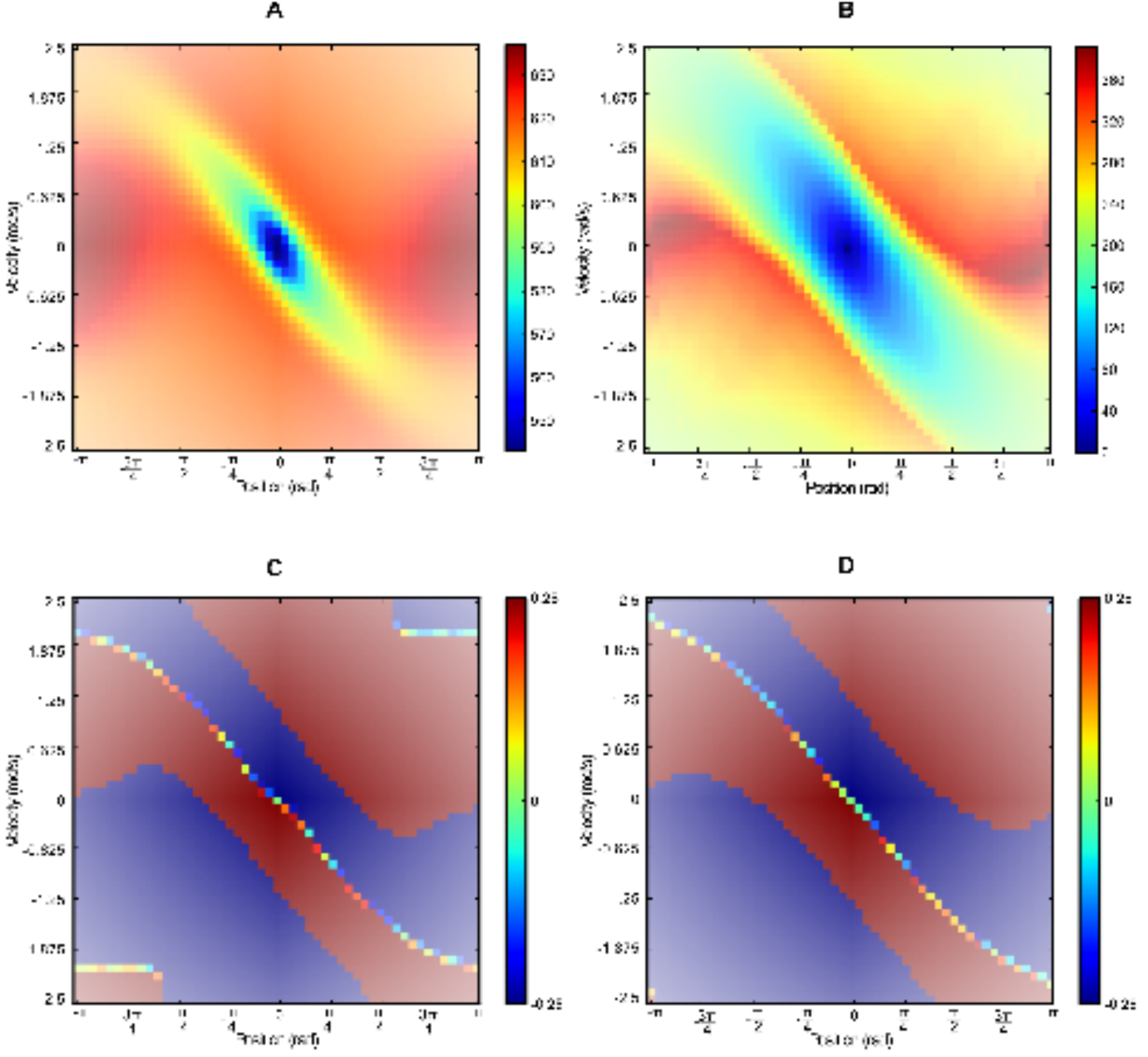}
\end{center}
\fi

\caption{{\bf Comparison of obtained value function, quasi-distance
    and policies for the inverted pendulum system.} (A) Value function obtained with undiscounted value
  iteration. (B) quasi-distance. (C) Policy obtained with undiscounted
  value iteration. (D) Policy
  obtained with the quasi-distance (most probable policy).}
\label{fig:pendul_bel_qd_all}
\end{figure}

Using the following cost function:
\[
g(x,u)=
\left\{
\begin{array}{l}
0\quad if\quad x=\text{goal and } u=0\\
1\quad \text{else}
\end{array} \right.
\]
the obtained Value function and quasi-distance are similar but not
equal (cf. Fig.~\ref{fig:pendul_bel_qd_all}).

For the Value function (without discount factor), the zero cost for
the goal state (needed for convergence) only propagates very slightly
and distant states have almost the same expected cost.

In contrast, the quasi-distance exhibits larger variations over states
because it is not smoothed by the computation of the mean cost
expectation of the Value Iteration method.

The constant cost chosen in this problem results in
minimizing ``path length'' (number of state transition) and
uncertainty (as the quasi-distance results in $\displaystyle \sum
\min_u \{ \frac{1}{p}\}$).

We computed the optimal policy with Value Iteration:
\[
\pi(x) = \argmin_u \left\{ g(x,u)+\sum_z \gamma V(z)p(z|x,u)
\right\}
\]

Similarly for the quasi-metric method we computed the argmin policy as the
policy minimizing the expected distance: 

\[
\pi(x) = \argmin_u \left\{ g(x,u)+\sum_zd(z,y)p(z|x,u)-d(x,y) \right\}
\]

Figure \ref{fig:pendul_bel_qd_all} shows the deterministic policies obtained for
both dynamic programming and quasi-distance. Here
again, small differences occur even though the policies are mostly
\emph{bang-bang}. 
We notice that small differences also occur due to the border
effect that is provoked by discretization. 

Despite these differences in both the Value function and the policy,
overall behavior is very similar.

\begin{figure}[ht]
\begin{center}

\iffig
\begin{center}
\includegraphics[width=4in]{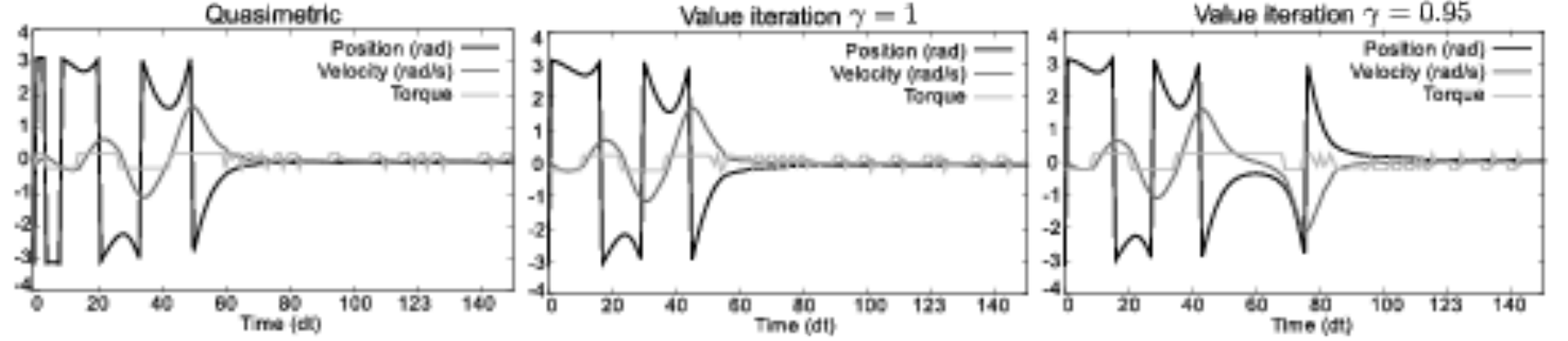}
\end{center}
\fi

\caption{
{\bf Behaviors obtained with quasimetric and dynamic
programming methods with different discount factors.}
Starting from the initial stable state, both methods lead to the
objective but with different trajectories.
}
\label{fig:pendul_pol_bel_qd_ctrl}
\end{center}
\end{figure}

Results of these policies can be seen on figure
\ref{fig:pendul_pol_bel_qd_ctrl}, starting from position $\pi$ with a velocity of
$0$. We can see that the trajectory obtained with the quasimetric
method is slightly longer than that obtained with undiscounted Value
iteration (optimal), but still better than that obtained with discounted Value
iteration (suboptimal with $\gamma=0.95$).

\begin{figure}[ht]
\iffig
\begin{center}
\includegraphics[width=4in]{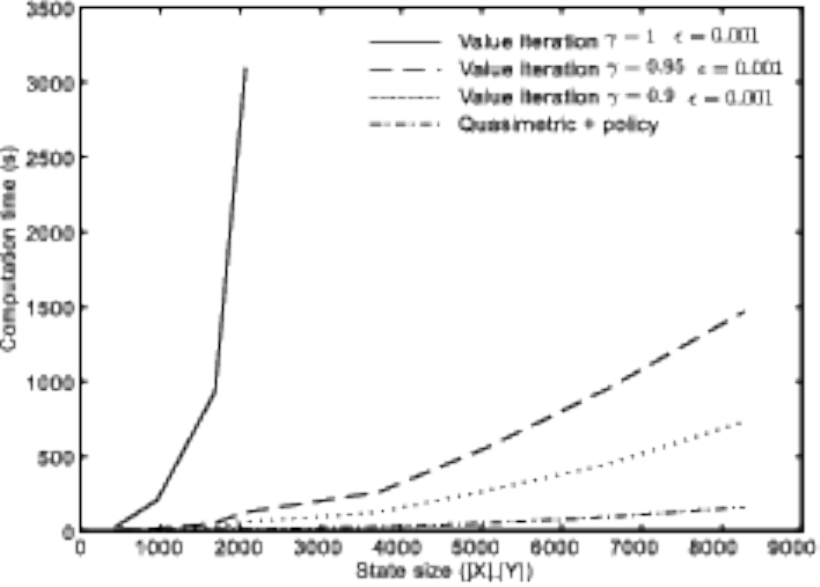}
\end{center}
\fi

\caption{{\bf Comparison of computation time for the under-actuated
  pendulum example.}}
\label{fig:pendul_timing_comp}
\end{figure}

We also compared computation time in terms of state space size
($|X|\times|Y|$ with a constant action space size $|U|=21$) between
discounted Value iteration and quasimetric methods.
Figure \ref{fig:pendul_timing_comp} shows the results obtained for the
single-goal quasi-distance (quasi-distance from all states to the goal) and its associated $P(U|X)$
policy, along with Value iteration with different discount factors.
These results were computed based on the same input transition
probabilities with single thread C++ implementations of algorithms on an
Intel Core2 Duo E6700 @ 2.66GHz desktop computer.

We can see that Value iteration heavily depends on the chosen
discount factor and is much slower than the quasimetric method for discounts close
to 1.  
Notice that computation time for the quasimetric method includes graph
construction (which should be done only once), quasi-distance and policy.
As an illustration, for a state space of
$|X|\times|Y|=91\times91=8281$, graph construction takes 129s,
quasi-distance (Dijkstra shortest path) takes 3s and policy takes 31s
while Value iteration with $\gamma=0.95$ takes 1469s.

\subsubsection*{Nonholonomic system}

A slightly more complicated system is the Dubins car model
\cite{Dubins1957}. This system is interesting because it exhibits
nonholonomic constraints for which optimal control is
difficult. However, it has generated a large amount of work during
last decades and several studies have provided in-depth
understanding and formal solutions of such systems and successfully
applied optimal methods for real-world robots (see
\cite{Laumond1998,Soueres1998}).

A Dubins car nonholonomic system is described with:
\begin{equation}\label{eq:nh}
\begin{pmatrix} \dot{x}_t \\ \dot{y}_t \\ \dot{\theta}_t \end{pmatrix} = 
\begin{pmatrix} cos \theta_t \\ sin \theta_t \\ 0 \end{pmatrix} u_l + 
\begin{pmatrix} 0 \\ 0 \\ 1 \end{pmatrix} u_a
\end{equation}

with the control input $u_l$ and $u_a$ respectively the linear and
angular velocity. 

For the sake of simplicity, we constrain the linear velocity to a constant value
$u_l=1.0$ and the angular velocity $u_a=u \in \left[-1;1 \right]$.

If we consider a probabilistic version of this system,
the transition probabilities for the dynamic model are:

\begin{equation}\label{eq:nh_fullprob}
P(X^{t+\Delta t},Y^{t+\Delta t},\Theta^{t+\Delta t}|X^{t},Y^{t},\Theta^{t},U^t)
\end{equation}

which can be rewritten with some independence assumptions as the product: 
\begin{equation}\label{eq:nh_indprob}
P(X^{t+\Delta t}|X^{t},\Theta^{t})P(Y^{t+\Delta t}|Y^{t},\Theta^{t})P(\Theta^{t+\Delta t}|\Theta^{t},U^t)
\end{equation}

and the following discrete Gaussian forms:

\begin{equation}\label{eq:nh_xytheta}
\left\{
\begin{array}{l l l}
P(X^{t+\Delta t}|[X^{t}=x],[\Theta^{t}=\theta])&\propto&\mathcal{N}(\mu_x=x+(cos(\theta)).\Delta t,\sigma_x)\\
P(Y^{t+\Delta t}|[Y^{t}=y],[\Theta^{t}=\theta])&\propto&\mathcal{N}(\mu_y=y+(sin(\theta)).\Delta t,\sigma_y)\\
P(\theta^{t+\Delta t}|[\Theta^{t}=\theta],[U^t=u])&\propto&\mathcal{N}(\mu_{\theta}=\theta+u.\Delta t,\sigma_{\theta})
\end{array} \right.
\end{equation}


We computed the quasimetric for this system in a discretized
state-space with the following parameters:

\[
X=[-5,5] \quad Y=[-5,5] \quad \Theta=[-\pi,\pi] \quad U=[-1,1]
\]

\[
|X|=|Y|=|\Theta|=51 \quad |U|=11
\]

\[
\sigma_x=\sigma_y=\sigma_{\theta}=0.05 \quad \Delta t=0.25
\]

\begin{figure}[ht]

\iffig
\begin{center}
\includegraphics[width=4in]{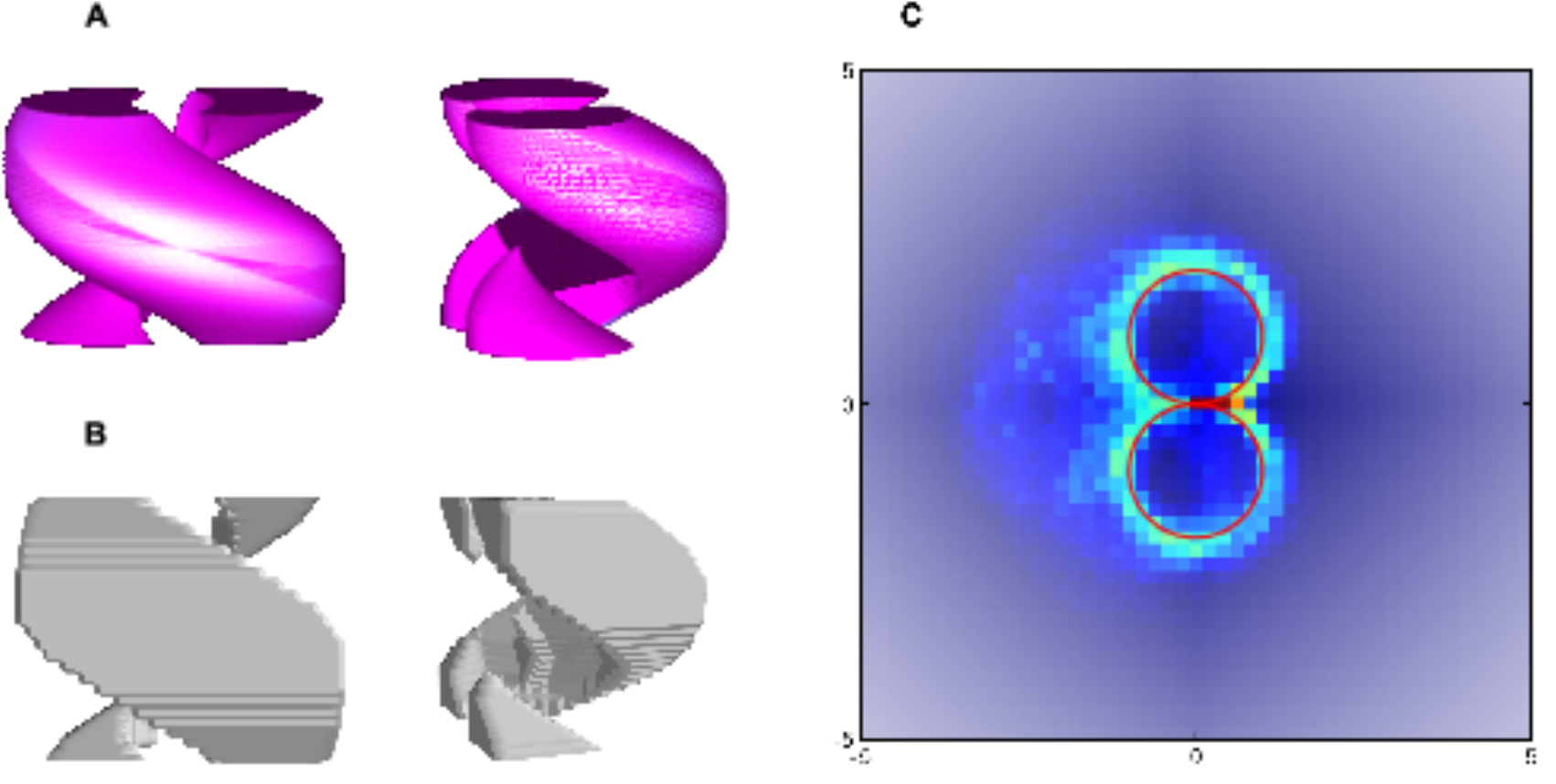}
\end{center}
\fi

\caption{{\bf Results obtained for the non-holonomic system.} (A) Accessibility volume of the Dubins car obtained with
  geometrical methods (adapted from \cite{Laumond1998}). (B) Discrete accessibility volume obtained
  for the described probabilistic case using the quasimetric method
  (axes aligned similarly).
  (C) Average trajectory for 500 simulations of 50 timesteps obtained
  starting at $(x,y,\theta)=(0,0,0)$ with goal at
  $(x,y,\theta)=(0,0,0)$ with a stochastic simulation and a drawn
  policy. Red curves are the optimal trajectories for a deterministic
  system.}
\label{fig:nonholo_all}
\end{figure}

The accessibility volume obtained from the origin -- i.e. the volume of the state space that
can be reached with a path length inferior to a given value -- is very
similar to the optimal deterministic one from \cite{Laumond1998}
(cf. Fig.~\ref{fig:nonholo_all}).
Although it is not clear what signification accessibility
volume could have for a probabilistic model, these two methods behave
quite similarly. 
It is to be noted that this result was obtained with our general
method without any specificity about the problem. 

If we simulate this system by adding noise to the position, according to
the described model, and use a random draw policy as described before we can see that the average trajectories obtained correspond mostly to direct loops (cf. Fig. \ref{fig:nonholo_all}).

These two symmetrical loops are comparable to the optimal deterministic
behavior. On average, the behavior of the system
closely match the theoretical deterministic optimal.
Due to the noise, it is almost impossible to
reach the goal in one trial. Most of the time the goal is
missed (with one trajectory passing very close) and then the controller
starts another loop or a figure of eight (trajectories on the left side
of the red curves). 

\section*{Conclusion}

We propose a new general method for control or decision making under
uncertainty. This method applies for the discrete MPD case with
positive cost and infinite or indefinite horizon.
The principle of this approach is to define a goal independent quasimetric to the
state space which can then be used to compute a policy for a chosen
goal or set of goals. Thus, each distances from all to one state describing a subspace
of the whole quasimetric can be viewed as an approximation of the
Value function.

To compute the distances between states we proposed to used the ``mean
cost per successful attempt'' of a direct transition that we
propagate by triangle inequality.
We show that this distance computation can be reformulated as a
standard deterministic shortest path problem allowing the use of
efficient algorithms.
Thanks to this property we have shown that the quasimetric approach
may lead to a very significant gain in terms of computational cost compared to
dynamic programming.
Illustrative examples were treated and have shown very good results.

We have demonstrated that for systems with possible prison states
(excluding the goal), the quasimetric can significantly differ from the optimal
solution when prisons are ``avoidable''. Moreover this method is still
able to produce a solution for problems with ``unavoidable'' prisons
where standard dynamic programming approach cannot. 
We proposed a way to finely tune risk sensitivity and
risk/cost trade-off, defining
risky states and a possible threshold on risk taking. 
Interestingly, this kind of risk-related behavior is reminiscent of that present in humans
and is still to be compared to classic methods in human decision making.

We also proposed a soft-max like way to compute a policy, which
provides an entire distribution rather than a unique optimal deterministic action.
Dealing with a probability distribution over actions provides, in our
sense, a less restrictive way of considering control under
uncertainty. With this method it is for instance, possible to make a
decision when faced with multiple equivalent actions, thus introducing
variability in actions and allowing exploration. This soft-max method,
along with the random draw action, is also applicable to the Value
function.

Extending this method to the HMM cases we described, is computationally very cheap
compared to the optimal POMDP, which is usually intractable. Moreover,
one can question whether solving the POMDP is relevant when the model
is imperfect or may change over time. 
Although our method is not optimal in the general
case and lacks information-gathering behavior, we think it could be a
useful bootstrap for learning using available prior knowledge,
even if the latter is very coarse.
Indeed, it could occasionally be more interesting to use a simple model with
uncertainties than a very complicated model which is nonetheless rarely perfect.  
Therefore, we could consider our method as a trade-off between solving the POMDP and
learning from scratch.

Finally, this very general approach can be applied to a wide range of
problems involving control under uncertainty. Although it is currently
restricted to discrete space, infinite/indefinite horizon cases, we hope to see
contributions from the community of control and planning as much of
the techniques developed for dynamic programming can be applied to
this method.

\section*{Acknowledgments}

We wish to express our gratitude to Daniel Bennequin and Julien Diard
for their very useful comments.

\bibliography{biblio}

\section*{Figure Legends}

\section*{Tables}

\begin{table}[ht]
\centering
\caption{Quasi-distances and Value function for example
  \ref{fig:exemples}A}
\label{tab:ex1_qd_v}
\begin{tabular}{|c|c|c|c|c|c|c|}
  \hline
  &$A$ &$B$ & $C$ & $D$ & $E$ & $V(.)$\\
  \hline
  $A$&0&3&4&4&5&4.5\\
 \hline
  $B$&$+\infty$ &0 &$+\infty$ &$+\infty$ &2 &2\\
 \hline
  $C$&$+\infty$ &$+\infty$ &0 &$+\infty$ &2.5 &2.5\\
 \hline
  $D$&$+\infty$ &$+\infty$ &$+\infty$ &0 &2.5 &2.5\\
 \hline
  $E$&$+\infty$ &$+\infty$ &$+\infty$ &$+\infty$ &0 &0\\
  \hline
\end{tabular}
\end{table}

\begin{table}[ht]
\centering
\caption{Quasi-distances and Value function for example
  \ref{fig:exemples}B}
\label{tab:exfrogger_qd_val}
\begin{tabular}{|c|c|c|c|c|c|c|}
  \hline
  &$A$ &$B$ & $C$ & $D$ & $V(.)$\\
  \hline
  $A$&0&$1$
  &$1+\frac{1}{\varepsilon}$&$\min\{1+\frac{1}{1-\varepsilon},\Omega\}$&$\Omega$\\
 \hline
  $B$& $+\infty$&0 &  $\frac{1}{\varepsilon}$&
 $\frac{1}{1-\varepsilon}$ & $+\infty$\\
 \hline
  $C$& $+\infty$ &  $+\infty$& 0& $+\infty$&  $+\infty$\\
 \hline
  $D$& $+\infty$ &  $+\infty$& $+\infty$&  0&0\\
  \hline
\end{tabular}
\end{table}

\end{document}